\documentclass[pdflatex,sn-mathphys-num]{sn-jnl}

\usepackage{graphicx}%
\usepackage{multirow}%
\usepackage{amsmath,amssymb,amsfonts}%
\usepackage{amsthm}%
\usepackage{mathrsfs}%
\usepackage[title]{appendix}%
\usepackage{xcolor}%
\usepackage{textcomp}%
\usepackage{manyfoot}%
\usepackage{booktabs}%
\usepackage{algorithm}%
\usepackage{algorithmicx}%
\usepackage{algpseudocode}%
\usepackage{listings}%
\usepackage{tabularx}
\usepackage{array}
\usepackage{caption}
\definecolor{paretogold}{HTML}{B8860B}

\theoremstyle{thmstyleone}%
\theoremstyle{thmstyletwo}%

\theoremstyle{thmstylethree}%

\newcommand{\paretofuse}{ParetoFuse}
\newcommand{\indepemb}{IndepEmb}
\newcommand{\rprox}{\ensuremath{\mathit{rProx}}}
\newcommand{\cmb}{\ensuremath{\mathit{CMB}}}
\newcommand{\pcov}{\ensuremath{\mathit{PCov}}}
\newcommand{\rspread}{\ensuremath{\mathit{R\text{-}spread}}}

\usepackage{multirow}

\begin{document}

\title[Article Title]{Searching the Pareto Front: Multi-Objective Retrieval for Heterogeneous Scientific Modalities}


\author*[1]{\fnm{Song Young} \sur{Oh}}\email{so27@uchicago.edu}

\author[2]{\fnm{Nesar} \sur{Ramachandra}}\email{nramachandra@anl.gov}

\author[3]{\fnm{Bin} \sur{Xia}}\email{xiabin@gatech.edu}

\author[4,1]{\fnm{Ian} \sur{Foster}}\email{foster@anl.gov}

\author[1,4]{\fnm{Kyle} \sur{Chard}}\email{chard@uchicago.edu}

\affil*[1]{\orgdiv{Department of Computer Science}, \orgname{University of Chicago}, \orgaddress{\street{5730 S Ellis Ave}, \city{Chicago}, \postcode{60637}, \state{IL}, \country{USA}}}

\affil[2]{\orgdiv{Computational Science Division}, \orgname{Argonne National Laboratory}, \orgaddress{\street{9700 S Cass Ave}, \city{Lemont}, \postcode{60439}, \state{IL}, \country{USA}}}

\affil[3]{\orgdiv{School of Physics}, \orgname{Georgia Institute of Technology}, \orgaddress{\street{837 State St NW}, \city{Atlanta}, \postcode{30332}, \state{GA}, \country{USA}}}

\affil[4]{\orgdiv{Data Science and Learning Division}, \orgname{Argonne National Laboratory}, \orgaddress{\street{9700 S Cass Ave}, \city{Lemont}, \postcode{60439}, \state{IL}, \country{USA}}}


\abstract{In cosmology, surveys span different sky regions and wavelength ranges, resulting in uneven modality coverage across observational datasets. Such heterogeneity can be addressed through either multimodal foundation models (FMs), which learn shared embedding spaces, or independently learned modality embeddings combined via late fusion. We investigate whether the latter provides an adaptive alternative to FM-based scientific retrieval. We evaluate this approach on a galaxy retrieval task spanning diverse modalities---history of star formation, spectra of light emitted, and multi-wavelength observations from two telescopes---comparing an FM baseline with separately trained embeddings fused through multi-objective optimization based on Pareto optimality. We find that the FM baseline achieves strong embedding-space proximity, but our method preserves modality-specific distance structure that allows scientists to inspect which physical signals drive similarity. Our method also achieves improved fidelity in raw-data space: across six measured properties, its nearest neighbor is 8\% closer to the query on average than that retrieved by kNN in the FM embedding space (spread 0.128 vs.\ 0.139). These results suggest that multi-objective retrieval can surface astrophysically coherent candidates that single-objective search alone may overlook.}




\maketitle

\section{Introduction}
\label{sec:intro}

Cross-modal retrieval is often conducted using nearest-neighbor search in a unified embedding space learned by a multimodal foundation model (FM)~\cite{lin2025mm, jia2021scaling}. In this paradigm, heterogeneous modalities are mapped to a shared representation, where retrieval can be performed with similarity metrics~\cite{wang2016comprehensive, wang2025cross}. This approach has been highly effective for domains in which modalities are well aligned and jointly available during training~\cite{radford2021learning, li2022blip, girdhar2023imagebind}. Scientific data, however, often violate these assumptions. Observations are incomplete, irregular, and shaped by modality-specific measurement processes. In astronomy, for example, telescope surveys differ in sky coverage, wavelength range, spatial resolution, and observational depth~\cite{tyson2002large, shanks2015vlt}. As a result, the same galaxy may be observed through only a subset of possible measurements, making globally aligned representations challenging to learn and difficult to maintain as new instruments or derived products become available.

Scientific retrieval is challenging because similarity may be defined differently across modalities. General-purpose FMs are designed to support broad downstream use, including missing-modality prediction and fine-tuning on limited data~\cite{li2024multimodalfm, bommasani2021opportunities}; however, retrieval imposes a more specific requirement: the induced neighborhood structure should preserve scientifically meaningful similarity relationships~\cite{wu2024scimmir}. This requirement is non-trivial to satisfy when modalities differ substantially in scale, noise, coverage and physical interpretation. Moreover, each modality may have its own embedding geometry, making direct comparison or aggregation across modalities poorly defined~\cite{jha2026harnessing, poklukar2022geometric}. Collapsing these modality-specific distances into a single distance measure can obscure scientifically relevant structure, while full retraining or re-alignment may be impractical when modalities evolve~\cite{xu2024survey}. These challenges call for retrieval methods that can operate over available modalities while preserving their distinct geometric contributions.

To address this gap, we propose \paretofuse, a multimodal retrieval framework that treats per-modality similarities as complementary signals rather than forcing them into a single global representation~\cite{fonseca1995overview, miettinen1999nonlinear}. Each modality produces its own distance estimate, and candidate objects are compared in the resulting multi-dimensional distance space. Pareto dominance is then used to identify candidates that are not uniformly worse than others across the available modalities~\cite{censor1977pareto}. Our design applies a lightweight late fusion mechanism for heterogeneous scientific data: it preserves modality-specific geometry while still producing a coherent candidate set for retrieval.

In this paper, we make the following contributions. We introduce a multi-objective formulation of heterogeneous scientific retrieval that retains modality-specific distances. We apply a Pareto-based reranking, including a post-processing step that selects precise neighbors from the non-dominated candidate set. We evaluate our proposed approach on a galaxy retrieval task against MOSAIC-lite, a lightweight variant of the MOSAIC multimodal FM~\cite{xia2025multi}, using both embedding-space and raw-data metrics, and find that our method preserves modality-specific distance structure while reducing average nearest neighbor spread in raw-data space by 8\%.

The remainder of this paper is organized as follows. Section~\ref{sec:related} reviews related work on multimodal representation learning and late fusion for retrieval. Section~\ref{sec:methods} presents \paretofuse, including independent embedding generation, one-dimensional distance projection, and multi-objective reranking methods. Section~\ref{sec:setup} describes the galaxy retrieval task, baselines, and evaluation metrics. Section~\ref{sec:results} reports the empirical comparison of FM retrieval and multi-objective retrieval using independent embeddings. Section~\ref{sec:discussion} discusses the implications for scientific retrieval, Section~\ref{sec:limits} summarizes limitations, and Section~\ref{sec:conclusion} concludes the paper. Appendix~\ref{secA1} provides additional details on MOSAIC and MOSAIC-lite.

\section{Related Work}
\label{sec:related}

\subsection{Multimodal Representation Learning}

Multimodal representation learning seeks to align heterogeneous data types in a shared embedding space~\cite{guo2019deep}. Contrastive learning has become a central paradigm for this goal, using paired observations to bring semantically related samples closer while separating unrelated ones~\cite{singh2023contrastive}. Large-scale models such as CLIP~\cite{radford2021learning} and ALIGN~\cite{jia2021scaling} demonstrated the effectiveness of this approach for vision--language alignment. Subsequent models have extended the paradigm through richer training objectives and broader modality coverage. For example, BLIP~\cite{li2022blip} combines contrastive and generative objectives, while ImageBind~\cite{girdhar2023imagebind} extends joint embedding learning across multiple sensory modalities. These systems establish a common retrieval design: modality-specific encoders map inputs into a shared latent space, and retrieval is then performed by nearest-neighbor search under a single similarity metric.

Scientific domains have increasingly adopted this shared-embedding paradigm, but under substantially different data conditions~\cite{leung2024towards}. In science, modalities often correspond to distinct instruments, measurement processes, spatial or temporal resolutions, and physical quantities. Recent work on omnimodal foundation models aims to address this heterogeneity by learning unified representations across diverse scientific signals~\cite{luo2025next, tyagi2025omnilearn}. In astronomy, models such as AION-1~\cite{parker2026aion}, AstroCLIP~\cite{parker2024astroclip} and MOSAIC~\cite{xia2025multi} learn joint embeddings for domain-specific modalities, supporting cross-modal retrieval at large dataset scale. These approaches are effective when modalities can be aligned through paired supervision. Our work instead considers a different setting in which joint training is not feasible, and retrieval must instead operate flexibly over independently learned modality representations.

\subsection{Late Fusion for Scientific Retrieval}

Late fusion provides a natural alternative when modalities are incomplete and independently maintained. Instead of learning a single joint representation, late-fusion methods compute modality-specific similarity scores and combine them at inference time~\cite{ye2012robust, zhao2024deep}. This separation between representation learning and decision-level integration is attractive in scientific workflows, where data sources may be distributed, updated asynchronously, or available only for subsets of samples~\cite{li2024multimodal, jin2025multimodal, oh2025smurf}. It also allows new modalities or revised encoders to be incorporated without retraining an entire multimodal model~\cite{lahat2015multimodal, atrey2010multimodal}.

A central challenge in late fusion is score heterogeneity. Similarity scores from different feature spaces may differ in scale, density, and discriminative behavior~\cite{snoek2005early}. Classical methods address this issue through normalization and weighted combinations, including linear score fusion and related ensemble methods~\cite{jain2005score, escalante2008late}. Other approaches model score distributions probabilistically~\cite{nandakumar2008likelihood}, learn supervised fusion weights~\cite{terrades2008optimal, shoukat2022late, gialampoukidis2016hybrid, hori2017attention}, or project modalities into shared latent subspaces~\cite{wright2009robust, dong2018late}.

Multi-objective optimization (MOO) offers a complementary view. Rather than resolving heterogeneous evidence through calibration, weighting, or projection, MOO treats each objective separately and identifies non-dominated trade-offs~\cite{fonseca1995overview, diwekar2003, marler2004survey, deb2011, giagkiozis2015methods}. A candidate is Pareto-optimal when no other candidate improves one objective without degrading another~\cite{miettinen1999nonlinear, qu2021comprehensive, gunantara2018review}. This principle has been applied in retrieval and recommendation settings that balance criteria such as relevance, diversity, and fairness~\cite{dai2011multi, arevalillo2013hybrid, lerher2013multi, van2016balancing, borovinvsek2017multi, rodriguez2020multi}. In scientific retrieval, this perspective is useful because disagreement across modalities may reflect meaningful physical variation rather than noise or calibration error. We adopt this approach for multimodal retrieval by preserving modality-specific distances as separate retrieval objectives. Candidates are then selected as Pareto-optimal trade-offs across modalities.

\section{Methods}
\label{sec:methods}

In this section, we first define the multimodal retrieval setting and notation of \paretofuse.  We then describe how modality-specific embeddings are learned independently, how these embeddings are converted into one-dimensional distance signals, and how candidates are ranked using multi-objective retrieval.

\subsection{Problem Formulation}
\label{sec:methods-problem}

We consider retrieval over objects with heterogeneous modality observations. Our goal is to rank corpus objects by their modality-specific similarity to a query, without reducing all modalities to a single shared embedding space.

Let $M$ denote the number of modalities. Each modality $m \in \{1,\ldots,M\}$ has an input space $\mathcal{X}^{(m)}$ and an encoder
\begin{equation}
f^{(m)} : \mathcal{X}^{(m)} \rightarrow \mathbb{R}^{d_m}.
\label{eq:encoder}
\end{equation}
The encoder maps observations from modality $m$ to a $d_m$-dimensional embedding space. An object $x$ may contain observations from several modalities and is written as $x=\left(x^{(1)},\ldots,x^{(M)}\right)$, where $x^{(m)} \in \mathcal{X}^{(m)}$ when modality $m$ is available. In scientific datasets, not every object is observed in every modality. We therefore let $\Omega(q,x) \subseteq \{1,\ldots,M\}$ denote the set of modalities available for both the query $q$ and corpus item $x$. For each available modality, we define a modality-specific distance 
\begin{equation}
\Delta_m(q,x)
=
\delta_m
\left(
f^{(m)}(q^{(m)}),
f^{(m)}(x^{(m)})
\right),
\quad m \in \Omega(q,x),
\label{eq:modality-distance}
\end{equation}
where $\delta_m$ is the distance function used in the embedding space of modality $m$. 
Smaller values of $\Delta_m(q,x)$ indicate greater similarity between $q$ and $x$ under modality $m$.

Rather than collapsing modality-specific distances into one global score, we treat each shared modality as a separate retrieval objective. For query $q$ and candidate $x$, we define the distance profile
\begin{equation}
\mathbf{\Delta}(q,x)
=
\left(
\Delta_m(q,x)
\right)_{m \in \Omega(q,x)}.
\label{eq:distance-profile}
\end{equation}
This vector records the candidate's distance to the query in each modality. Given a retrieval corpus $\mathcal{C}$, the goal is to rank candidates $x \in \mathcal{C}$ using these modality-specific distance profiles.

\subsection{Independent Modality Embeddings}
\label{sec:methods-encoders}

\begin{table}[b]
\centering
\begin{minipage}{\columnwidth}
\captionsetup{width=\linewidth}
\caption{Summary of the four autoencoder modalities. Encoder structures are fully connected neural networks; dimensions denote layer widths from input to bottleneck. The first dimension corresponds to the dimensionality of each modality.}
\label{tab:autoencoders}
\footnotesize
\setlength{\tabcolsep}{4pt}
\renewcommand{\arraystretch}{1.15}
\begin{tabularx}{\linewidth}{@{}lXl@{}}
\toprule
\textbf{Modality} & \textbf{Brief description} & \textbf{Encoder structure} \\
\midrule
SPHEREx ($\mathbf{f}_{\rm sph}$) 
& Narrowband photometry across 102 spectral bandpasses 
& $102 \to 128 \to 64$ \\

LSST ($\mathbf{f}_{\rm lsst}$) 
& Broadband filter magnitudes ($ugrizy$) 
& $6 \to 16 \to 32 \to 16 \to 4$ \\

SFH ($\boldsymbol{\psi}$) 
& Stellar mass formed per time bin over 117 age bins 
& $117 \to 128 \to 64$ \\

SED ($\mathbf{f}_{\lambda}$) 
& Spectral energy distribution sampled at 921 wavelengths 
& $921 \to 512 \to 256$ \\
\bottomrule
\end{tabularx}
\end{minipage}
\end{table}

To obtain independent modality embeddings, we learn a separate representation per modality. For each modality $m$, we train an autoencoder with encoder $f^{(m)}$ and decoder $h^{(m)}$~\cite{rumelhart1986learning, hinton2006reducing}. The decoder is used only during training; after training, we use the bottleneck representation $f^{(m)}(x^{(m)})$ as the modality-specific embedding. \autoref{tab:autoencoders} summarizes the four modalities and encoder architectures we use in this work.

The reconstruction objective for modality $m$ is
\begin{equation}
\mathcal{L}^{(m)}_{\mathrm{AE}}
=
\frac{1}{N_m}
\sum_{i=1}^{N_m}
\ell_{\beta}
\left(
x_i^{(m)},
h^{(m)}(f^{(m)}(x_i^{(m)}))
\right),
\label{eq:autoencoder-loss}
\end{equation}
where $N_m$ is the number of training samples for modality $m$, and $\ell_{\beta}$ is the SmoothL1 loss with $\beta=1$. SmoothL1 is used instead of mean-squared error because it is less affected by extreme residuals, making training more stable across heterogeneous modalities.

\begin{figure}[t]
\centering
\includegraphics[width=0.7\textwidth]{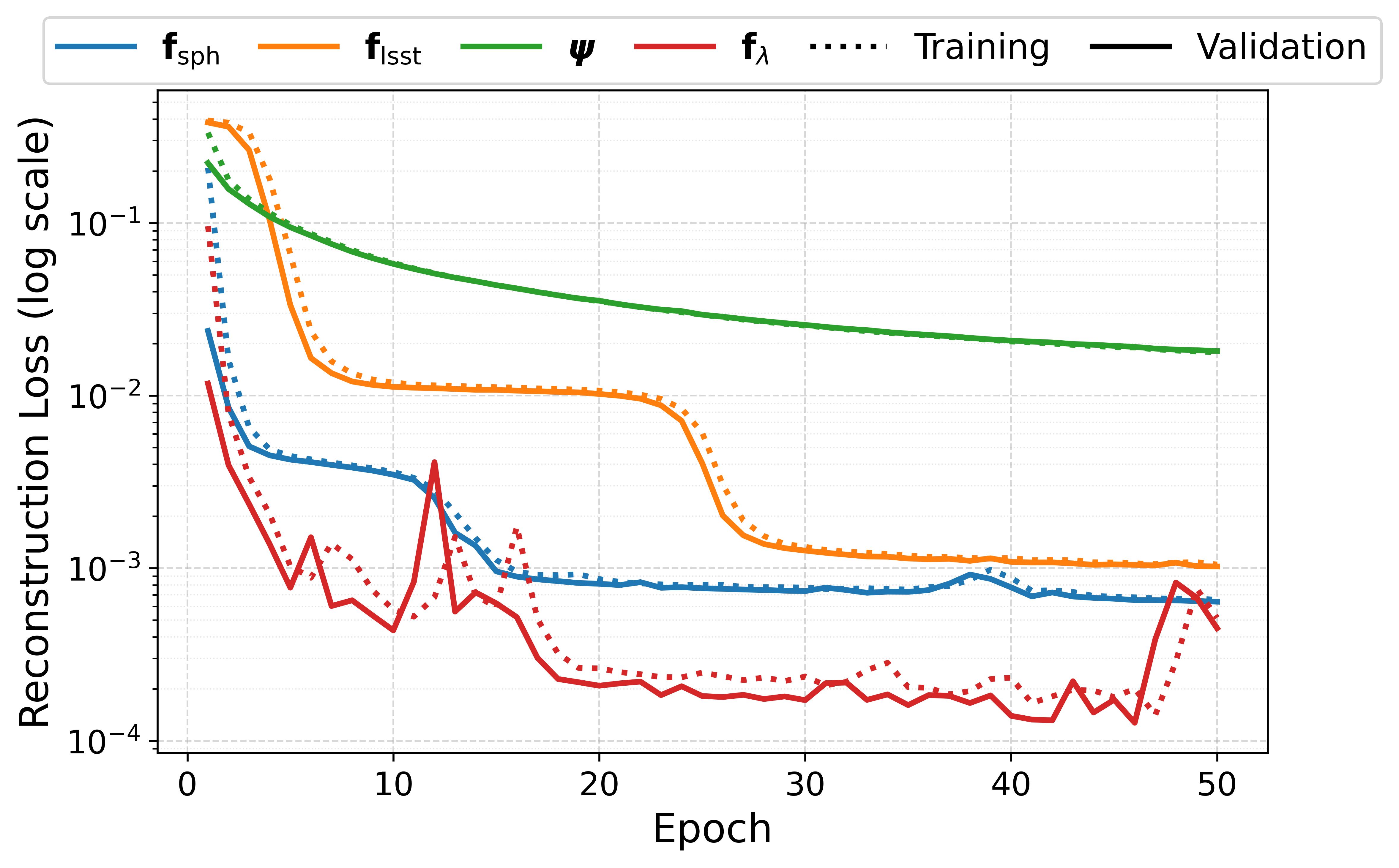}
\caption{Training and validation reconstruction losses for the independently trained autoencoders. Each model is trained only on its corresponding modality, and the bottleneck representation is used as the modality-specific embedding.}
\label{fig:loss_curves}
\end{figure}

Each decoder mirrors its corresponding encoder. Hidden layers use ReLU activations; bottleneck and output layers are linear. We train all models with Adam, using learning rate $10^{-3}$, batch size 1024, and 50 epochs, and retain the checkpoint with the lowest validation loss for each modality. The resulting training and validation reconstruction curves are shown in \autoref{fig:loss_curves}. The loss corresponding to star formation history shows the least optimization because it is less informative compared to other modalities, and has considerably lower signal-to-noise ratio. While the hyper-parameters and architectures of the individual embedding networks are tuned by-hand, automated searches can be utilized as well.

\begin{figure}[b]
\centering
\includegraphics[width=0.8\textwidth]{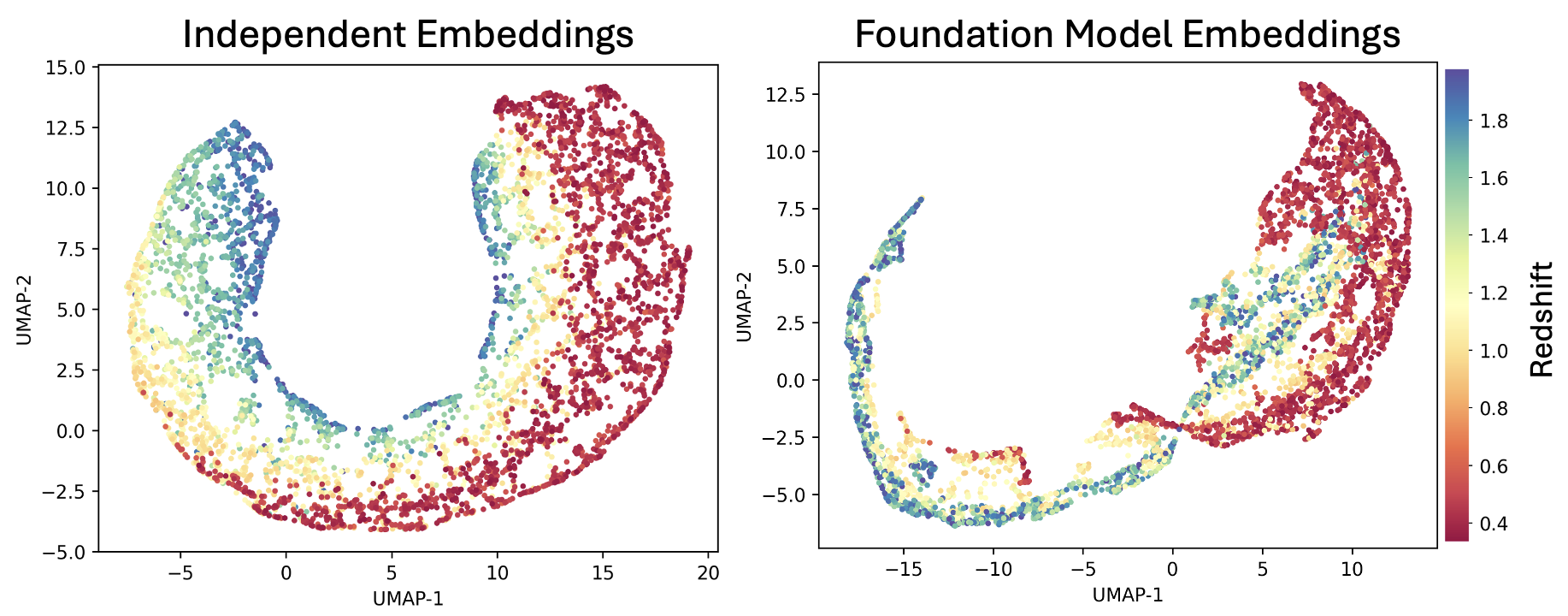}
\caption{UMAP visualization of $\mathbf{f}_{\rm lsst}$ embeddings for 5{,}000 galaxy objects. Left: independently trained embeddings. Right: foundation model embeddings. Points are colored by redshift values.}
\label{fig:umap}
\end{figure}

To qualitatively compare the latent geometries induced by independent modality training and the shared FM, \autoref{fig:umap} visualizes the corresponding $\mathbf{f}_{\rm lsst}$ embedding spaces. The distinct UMAP structures suggest that the two representations organize the data differently, although the projected coordinates and interpoint distances do not have direct physical interpretations.

\subsection{Projection to One-Dimensional Signals}
\label{sec:methods-projection}

We reduce each modality embedding to a one-dimensional representation, allowing embeddings of different dimensionalities to serve as comparable retrieval objectives. Query--candidate distances are then computed separately for each modality, yielding one scalar distance per modality while preserving them as distinct objectives for subsequent multi-objective reranking. For modality $m$, let
\begin{equation}
g^{(m)} : \mathbb{R}^{d_m} \rightarrow \mathbb{R}
\label{eq:projector}
\end{equation}
denote a modality-specific projector. The projected coordinate of item $x$ is
\begin{equation}
z_m(x)
=
g^{(m)}
\left(
f^{(m)}(x^{(m)})
\right),
\label{eq:projected-coordinate}
\end{equation}
and the corresponding query--candidate distance is
\begin{equation}
d_m(q,x)
=
\left|
z_m(q) - z_m(x)
\right|.
\label{eq:projected-distance}
\end{equation}
The projected distance profile used for reranking is
\begin{equation}
\mathbf{d}(q,x)
=
\left(
d_m(q,x)
\right)_{m \in \Omega(q,x)}.
\label{eq:projected-distance-vector}
\end{equation}

In this work, we evaluate two projectors with different modeling assumptions. The first is a neural projector, defined as a nonlinear mapping implemented by a multilayer perceptron with hidden dimensions 128 and 64. It is trained independently for each modality to preserve pairwise distances from the original embedding space after projection. The second is a PCA projector, defined by the leading principal component of each modality embedding space, and provides a linear baseline. This comparison assesses whether the retrieval stage benefits from nonlinear projection or whether a linear dominant-direction projection is sufficient.

\subsection{Multi-Objective Reranking}
\label{sec:methods-moo}

Multi-objective reranking retains modality-specific similarity information during candidate selection. Because different modalities may encode complementary structure, converting them into a single score can obscure meaningful trade-offs among candidates. We therefore rerank candidates using their projected distance profiles, treating each modality as a separate objective.

Given a query $q$, each candidate $x \in \mathcal{C}$ is represented by the projected distance vector $\mathbf{d}(q,x)$. We evaluate two reranking families: Pareto-based methods, which preserve the vector-valued distance profile, and scalarized methods, which reduce it to a scalar value.

\subsubsection{Pareto Reranking}
\label{sec:methods-pareto}

Pareto reranking identifies candidates whose modality-wise distance profiles are not dominated by any other candidate~\cite{fonseca1995overview, miettinen1999nonlinear, censor1977pareto}. For a query $q$, candidate $x_a$ Pareto-dominates candidate $x_b$, written as $x_a \prec_q x_b$, if
\begin{equation}
d_m(q,x_a)
\le
d_m(q,x_b)
\quad
\forall m \in \Omega(q,x_a,x_b),
\label{eq:pareto-weak}
\end{equation}
and there exists at least one modality $m^\star \in \Omega(q,x_a,x_b)$ such that
\begin{equation}
d_{m^\star}(q,x_a)
<
d_{m^\star}(q,x_b).
\label{eq:pareto-strict}
\end{equation}
Here, $\Omega(q,x_a,x_b)$ denotes the modalities available for the query and both candidates. Thus, $x_a$ dominates $x_b$ when it is no farther from the query in every modality and strictly closer in at least one. The Pareto front for query $q$ is the set of non-dominated candidates:
\begin{equation}
\mathcal{P}(q)
=
\left\{
x \in \mathcal{C}
:
\nexists x' \in \mathcal{C}
\text{ such that }
x' \prec_q x
\right\}.
\label{eq:pareto-front}
\end{equation}

\begin{figure}[t]
\centering
\includegraphics[width=0.7\textwidth]{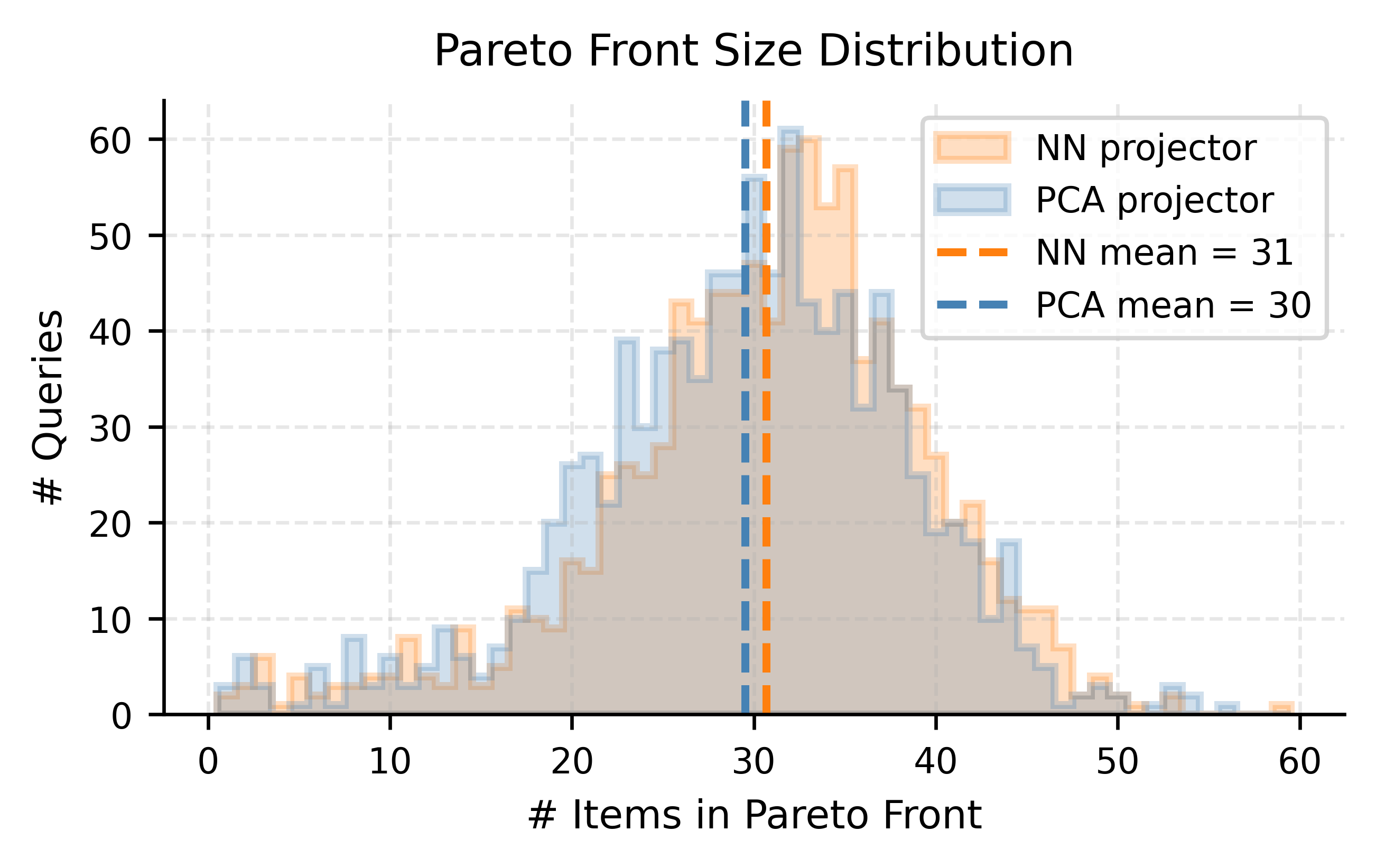}
\caption{Distribution of Pareto front sizes across 1{,}000 queries over a corpus of 5{,}000 items. Front sizes are computed in the four-dimensional projected distance space.}
\label{fig:pareto-front-size}
\end{figure}

The resulting front contains candidates for which no other candidate provides an unambiguous improvement across the available modalities. As shown in \autoref{fig:pareto-front-size}, the size of this non-dominated candidate set varies across queries.

\subsubsection{Scalarized Reranking}
\label{sec:methods-scalarization}

We include scalarized reranking as a comparison to Pareto retrieval. Scalarized methods reduce the projected distance vector $\mathbf{d}(q,x)$ to a single score and rank candidates by increasing score. To reduce scale effects, we use normalized distances $\tilde{d}_m(q,x)$, obtained by dividing $d_m(q,x)$ by the empirical mean distance for modality $m$ over the evaluation corpus. We consider three methods:

\vspace{0.35em}
\noindent\textbf{(1) Weighted sum.}
This method ranks candidates by
\begin{equation}
s^{\mathrm{wsum}}(q,x)
=
\sum_{m \in \Omega(q,x)}
\tilde{d}_m(q,x),
\label{eq:weighted-sum}
\end{equation}
favoring candidates with small average distance across modalities~\cite{marler2010weighted}.

\vspace{0.35em}
\noindent\textbf{(2) Chebyshev.}
This method ranks candidates by
\begin{equation}
s^{\mathrm{cheb}}(q,x)
=
\max_{m \in \Omega(q,x)}
\tilde{d}_m(q,x),
\label{eq:chebyshev}
\end{equation}
penalizing the largest modality-specific mismatch~\cite{boyd01}.

\vspace{0.35em}
\noindent\textbf{(3) $p$-norm.}
This method ranks candidates by
\begin{equation}
s^{p\text{-norm}}(q,x)
=
\left(
\sum_{m \in \Omega(q,x)}
\tilde{d}_m(q,x)^p
\right)^{1/p}.
\label{eq:p-norm}
\end{equation}
with $p=2$ in our experiments. Larger $p$ values make the score more sensitive to large modality-specific mismatches~\cite{gentile1999robustness}.

\subsection{Pareto Gold Refinement}
\label{sec:methods-gold}

The Pareto front has query-dependent size and therefore does not directly produce a fixed-$K$ retrieval list. We define \emph{Pareto Gold} as a two-stage refinement: first compute the Pareto front $\mathcal{P}(q)$, then rank only the non-dominated candidates by nearest-neighbor distance in the concatenated independent embedding space.

\begin{figure}[t]
\centering
\includegraphics[width=\textwidth]{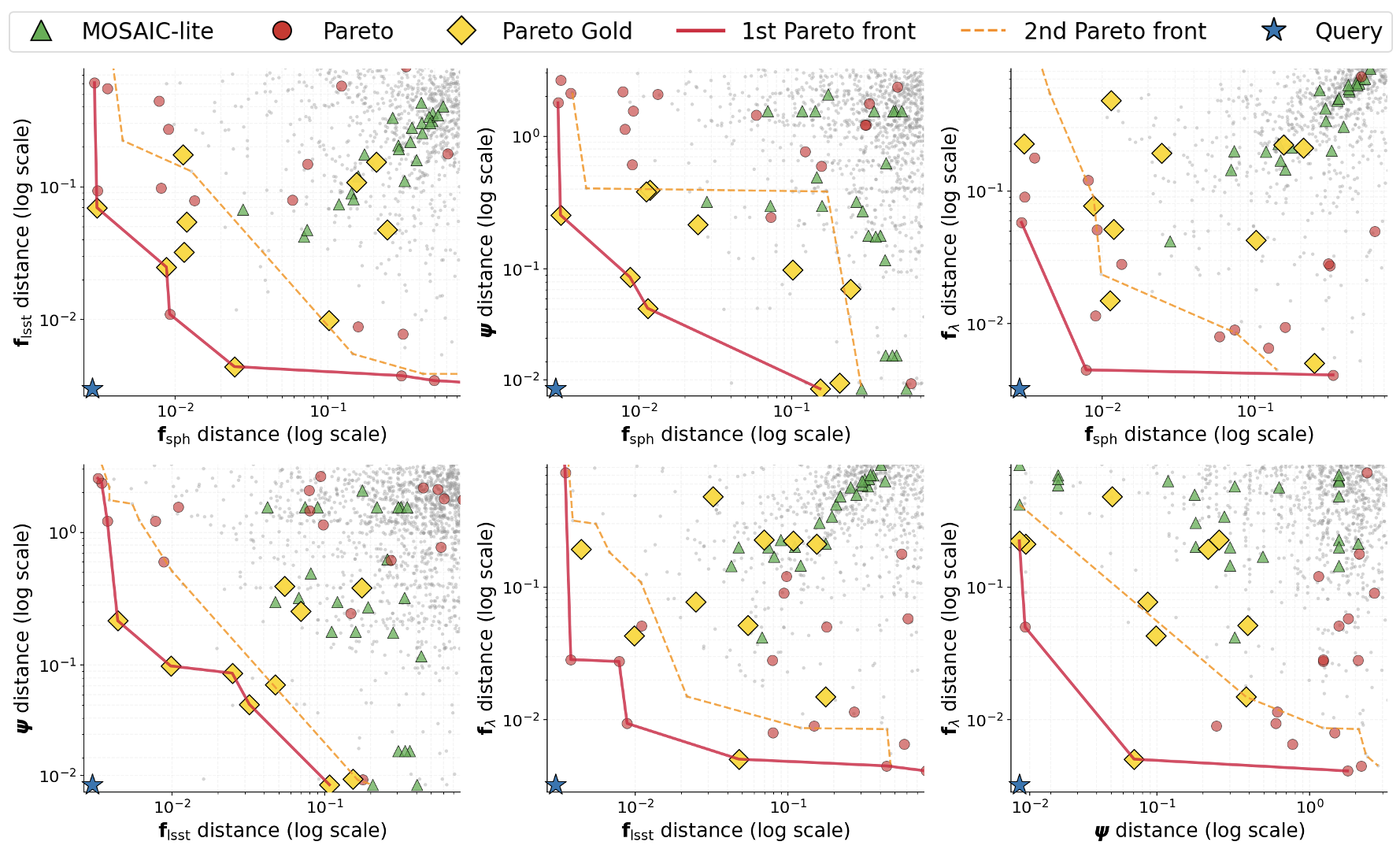}
\caption{Pairwise 2D views of modality-specific distances from a query galaxy across four modalities: $\mathbf{f}_{\rm sph}$, $\mathbf{f}_{\rm lsst}$, $\boldsymbol{\psi}$, and $\mathbf{f}_{\lambda}$. Each axis shows a log-scaled distance in the corresponding modality units. Each panel shows one modality pair. Gray points are corpus items, and the query appears at the origin because its self-distance is zero. Red circles indicate items on the first Pareto front, while the dashed orange curve shows the second Pareto front for reference. Green triangles denote multimodal FM $k$NN results, whereas yellow diamonds denote results from the \emph{Pareto Gold} refinement of \paretofuse, obtained by applying $k$NN ($k=10$) within the Pareto set. Because Pareto retrieval is computed in the full 4D distance space, Pareto-optimal items need not form a visible front in 2D projection.}
\label{fig:pareto_projection}
\end{figure}

For fixed-size output, \emph{Pareto Gold} returns
\begin{equation}
\mathcal{R}^{\mathrm{PG}}_K(q)
=
\operatorname*{TopK}_{x \in \mathcal{P}(q)}
\left(
-\left\|
f_{\oplus}(q)-f_{\oplus}(x)
\right\|_2
\right),
\label{eq:pareto-gold}
\end{equation}
where $f_{\oplus}(\cdot)$ denotes the concatenation of the modality-specific embeddings. Thus, candidates must first be non-dominated across projected modality distances and are then ordered by local proximity in the concatenated independent embedding space. \autoref{fig:pareto_projection} illustrates this refinement for an example query, contrasting the full Pareto front with the resulting fixed-size retrieval set.

\section{Experimental Design}
\label{sec:setup}

This section describes the evaluation dataset, retrieval protocol, methods under comparison, and metrics used to assess retrieval quality.

\subsection{Dataset}
\label{sec:setup-data}

We evaluate our method using a galaxy retrieval corpus built from the Last Journey simulation~\cite{heitmann2021last}. The corpus contains $|\mathcal{C}|=5{,}000$ galaxies, and the query set contains $|\mathcal{Q}|=1{,}000$ held-out galaxies disjoint from the corpus. Each query is evaluated against the same corpus, enabling paired comparisons across retrieval methods.

Every galaxy contains four vector modalities, summarized in \autoref{tab:autoencoders}. These modalities differ in dimensionality and physical meaning, making them a suitable testbed for heterogeneous scientific retrieval. We also use redshift and stellar mass as scalar attributes for raw-space evaluation, as described later in \autoref{sec:setup-metrics}.

\subsection{Retrieval Protocol}
\label{sec:setup-protocol}

All comparison methods use the same candidate generation protocol. For each query $q$, we retrieve the top-$k$ nearest neighbors in each modality using the projected distance $d_m(q,x)$, then take the union of these modality-specific neighborhoods as the reranking pool. This preserves candidates that are close in at least one modality before applying cross-modal reranking.

We use $k=50$ for candidate generation and report the main results at retrieval depth $K=25$. The candidate depth is larger than the reporting depth so that reranking can select from a sufficiently broad pool while keeping computation fixed across methods. The main depth $K=25$ provides a moderate top-list size for analyzing neighborhood quality and is not tuned separately for any method. For overlap analyses, we also evaluate $K \in \{1, 5, 10, 25\}$ to characterize sensitivity to retrieval depth.

\subsection{Comparison Methods}
\label{sec:setup-baselines}

We compare late fusion methods over independently trained embeddings against standard $k$NN retrieval in a shared multimodal FM embedding space.

\vspace{0.35em}
\noindent\textbf{Independent-embedding methods.}
These methods use the independently trained embeddings described in \autoref{sec:methods-encoders}, denoted as \indepemb\ in the result table. We evaluate: 
\textbf{(i) Pareto fusion}, with both neural and PCA projectors, reported as Pareto (NN) and Pareto (PCA); 
\textbf{(ii) Scalar fusion}, including weighted-sum, Chebyshev, and $p$-norm scalarization using the neural projector; and 
\textbf{(iii) $k$NN}, which performs nearest-neighbor search over concatenated independent embeddings.

\vspace{0.35em}
\noindent\textbf{Shared-embedding FM baseline.}
As a shared-embedding baseline, we use MOSAIC-lite, a lightweight multimodal FM trained at the same data scale as our independent encoders. Retrieval is performed using $k$NN search in its unified embedding space.

\subsection{Evaluation Metrics}
\label{sec:setup-metrics}

Because multimodal galaxy retrieval has no canonical ground-truth ranking, we evaluate retrieval quality using query-relative diagnostics that assess retrieved items against a corpus-level reference. The four metrics capture different aspects of retrieval quality: \emph{relative proximity} (\rprox) measures overall closeness in the projected embedding space, \emph{cross-modal balance} (\cmb) measures how evenly that closeness is distributed across modalities, \emph{Pareto coverage} (\pcov) measures whether the retrieved list preserves non-dominated trade-offs, and \emph{raw-data spread} (\rspread) measures fidelity in the original raw-feature space.

Let $\mathcal{R}_K(q)\subset\mathcal{C}\setminus\{q\}$ be the top-$K$ items returned for query $q\in\mathcal{Q}$. For modality $m$, we define the corpus reference distance as

\begin{equation}
\bar{d}_m(q) = \frac{1}{|\mathcal{C}|-1} \sum_{x\in\mathcal{C}\setminus\{q\}} d_m(q,x).
\label{eq:corpus-mean}
\end{equation}

This reference normalizes distances by the average corpus distance for the same query and modality, making scores comparable across queries and modalities. All metrics are computed per query and then averaged over $\mathcal{Q}$.

\subsubsection{Relative proximity (\rprox)}
\label{sec:setup-rprox}

Relative proximity measures the overall compactness of the retrieved neighborhood in the projected embedding space. It asks whether the returned items are, on average, closer to the query than a typical corpus item. For each modality $m$,
\begin{equation}
\rprox_m(q)
=
\frac{1}{K}
\sum_{x\in\mathcal{R}_K(q)}
\frac{d_m(q,x)}{\bar{d}_m(q)},
\qquad
\rprox(q)
=
\frac{1}{|\Omega_K(q)|}
\sum_{m\in\Omega_K(q)}
\rprox_m(q),
\label{eq:rprox}
\end{equation}
where
\begin{equation}
\Omega_K(q)
=
\bigcap_{x\in\mathcal{R}_K(q)}
\Omega(q,x)
\label{eq:rprox-modalities}
\end{equation}
is the set of modalities shared by the query and all retrieved items. Values below 1 indicate retrievals closer than the corpus average; lower is better.

\subsubsection{Cross-modal balance (\cmb)}
\label{sec:setup-cmb}

Relative proximity (\rprox) summarizes average closeness, but it does not indicate whether that closeness comes from all modalities or only a subset of them. Cross-modal balance therefore measures how evenly retrieval quality is distributed across modalities. We compute the per-query coefficient of variation of the per-modality relative proximities,
\begin{equation}
\operatorname{cv}_{\rprox}(q)
=
\frac{
\operatorname{std}_{m\in\Omega_K(q)}\!\bigl(\rprox_m(q)\bigr)
}{
\operatorname{mean}_{m\in\Omega_K(q)}\!\bigl(\rprox_m(q)\bigr)
},
\label{eq:rprox-cv}
\end{equation}
and define cross-modal balance as
\begin{equation}
\cmb
=
1 \,-\, \frac{1}{|\mathcal{Q}|}
\sum_{q\in\mathcal{Q}}
\operatorname{cv}_{\rprox}(q).
\label{eq:cmb}
\end{equation}
Thus $\cmb=1$ when all per-modality \rprox\ values are identical, and \cmb\ decreases as the retrieved neighborhood becomes more uneven across modalities.

\subsubsection{Pareto coverage (\pcov)}
\label{sec:setup-pcov}

We use Pareto coverage to assess how many candidates in the top-$K$ list remain non-dominated relative to the other retrieved items. This metric quantifies how much non-dominated structure is preserved after retrieval. Let $\mathcal{P}_K(q)\subseteq\mathcal{R}_K(q)$ be the retrieved items that are not Pareto-dominated by any other item in $\mathcal{R}_K(q)$ under the dominance
relation $\prec_q$ from Eqs.~(\ref{eq:pareto-weak})--(\ref{eq:pareto-strict}).
Pareto coverage is
\begin{equation}
\pcov@K(q)
=
\frac{|\mathcal{P}_K(q)|}{K}.
\label{eq:pcov}
\end{equation}
In our experiments, we truncate each retrieved list at $K=10$. Higher values mean that a larger fraction of the top-$K$ list remains non-dominated. Unlike the corpus-level front $\mathcal{P}(q)$ in Eq.~(\ref{eq:pareto-front}), $\mathcal{P}_K(q)$ is computed only within the retrieved set; \pcov\ therefore characterizes the ranking rather than the full
corpus structure.

\subsubsection{Raw-data spread (\rspread)}
\label{sec:setup-rspread}

The preceding metrics are computed in the embedding space used for retrieval. R-spread provides a complementary check in the original feature space, measuring whether retrieved galaxies remain close to the query in raw physical units. For each modality $m$, we use
\begin{equation}
d^{\mathrm{raw}}_m(q,x)
=
\begin{cases}
\|q^{(m)}-x^{(m)}\|_2,
& m\text{ is vector-valued}, \\[2pt]
|q^{(m)}-x^{(m)}|,
& m\text{ is scalar}.
\end{cases}
\label{eq:raw-distance}
\end{equation}
Let $\bar{d}^{\mathrm{raw}}_m(q)$ be defined as in
Eq.~(\ref{eq:corpus-mean}) using $d^{\mathrm{raw}}_m$. We define
\begin{equation}
\rspread_m(q)
=
\frac{1}{K}
\sum_{x\in\mathcal{R}_K(q)}
\frac{d^{\mathrm{raw}}_m(q,x)}{\bar{d}^{\mathrm{raw}}_m(q)}.
\label{eq:rspread}
\end{equation}
Lower values indicate closer retrievals in the original feature space. We report \rspread\ for the four vector modalities (described in \autoref{tab:autoencoders}) and for two scalar modalities, redshift and stellar mass, which are not used to generate independent embeddings.

\section{Results}
\label{sec:results}

We organize the results around how each method shapes retrieved neighborhoods, preserves modality-specific trade-offs, and behaves in raw physical space.

\subsection{Embedding-space proximity favors shared geometry}
\label{sec:res-quality}

\begin{table}[b]
\centering
\caption{Retrieval performance under embedding-space metrics.}
\label{tab:main}
\footnotesize
\renewcommand{\arraystretch}{1.15}
\begin{tabular*}{0.8\columnwidth}{@{\extracolsep{\fill}}lccc@{}}
\toprule
\textbf{Method} & \textbf{\rprox} & \textbf{\cmb} & \textbf{\pcov} \\
\midrule
\textcolor{paretogold}{\textbf{Pareto Gold [\indepemb]}}
  & \textcolor{paretogold}{0.249}
  & \textcolor{paretogold}{0.044}
  & \textcolor{paretogold}{\textbf{0.480}} \\
Pareto (NN) [\indepemb]     & 0.346 & 0.109 & 0.440 \\
Pareto (PCA) [\indepemb]    & 0.348 & 0.115 & 0.448 \\
Scalar:wsum [\indepemb]     & 0.295 & 0.089 & 0.388 \\
Scalar:cheb [\indepemb]     & 0.293 & 0.068 & 0.364 \\
Scalar:pnorm [\indepemb]    & 0.299 & 0.109 & 0.381 \\
$k$NN [\indepemb]           & 0.247 & 0.291 & 0.417 \\
\midrule
$k$NN [Foundation Model]    & \textbf{0.153} & \textbf{0.810} & 0.382 \\
\bottomrule
\end{tabular*}

\vspace{0.35em}
\parbox{0.8\columnwidth}{\footnotesize
\textit{Note.} Lower \rprox\ is better: values below 1 mean that retrieved items are closer to the query object than a randomly drawn corpus item, measured in the method's own embedding space. Higher \cmb\ indicates more balanced per-modality proximity, but is not a universal optimality criterion. Higher \pcov\ indicates greater coverage of non-dominated items in the top-10 results.
\emph{Pareto Gold} is reported at $K{=}10$ to match the \pcov\ truncation. Bold marks the best value per column.
}
\end{table}

Under an embedding-space proximity criterion, the shared FM geometry provides the strongest retrieval neighborhoods. As summarized in \autoref{tab:main}, the FM baseline achieves the lowest \rprox\ (0.153), corresponding to a 38\% relative reduction compared with the strongest independent-embedding baseline, $k$NN [\indepemb] (0.247), and larger reductions compared with Pareto and scalar fusion methods. This result is expected: the FM embedding space maps the four modalities into a single 384-dimensional representation, so Euclidean $k$NN directly optimizes neighborhoods in the same geometry learned by the model.

Independent-embedding methods are less competitive under \rprox\ because they do not optimize a single shared distance function. Late fusion instead ranks candidates using modality-specific distance signals, which can surface items that are strong along particular modality axes but not uniformly close under the averaged proximity measured by \rprox.

\begin{table}[b]
\centering
\caption{Comparison of \emph{Pareto Gold} and the FM baseline.}
\label{tab:gold-recall}
\footnotesize
\renewcommand{\arraystretch}{1.15}
\begin{tabular*}{0.7\columnwidth}{@{\extracolsep{\fill}}ccc@{}}
\toprule
\textbf{Depth $K$}
& \textbf{$P(\geq 1$ overlap$)$}
& \textbf{Pareto Gold recall} \\
\midrule
$1$  & $0.556 \pm 0.031$ & $0.556 \pm 0.031$ \\
$5$  & $0.725 \pm 0.028$ & $0.183 \pm 0.028$ \\
$10$ & $0.823 \pm 0.024$ & $0.137 \pm 0.024$ \\
$25$ & $0.942 \pm 0.014$ & $0.110 \pm 0.015$ \\
\bottomrule
\end{tabular*}

\vspace{0.35em}
\parbox{0.7\columnwidth}{\footnotesize
\textit{Note.} Coverage probability reports whether the two methods share at least one retrieved item at depth $K$; recall measures the item-level fraction of \emph{Pareto Gold} results recovered by the FM. Intervals are 95\% confidence intervals over 1{,}000 queries.
}
\end{table}

\subsection{Multi-objective retrieval preserves modality-specific structure}
\label{sec:res-coverage}

We next examine whether the retrieved neighborhoods retain modality-specific trade-offs. The \cmb\ results exhibit a clear contrast between shared and independent representations. The FM baseline obtains the highest \cmb\ (0.810) in \autoref{tab:main}, whereas all independent-embedding methods are substantially lower: Pareto (NN) and Pareto (PCA) achieve 0.109 and 0.115, scalar methods range from 0.068 to 0.109, and $k$NN [\indepemb] reaches 0.291. Since lower \cmb\ reflects greater variation among per-modality proximity values, these results indicate that multi-objective retrieval preserves heterogeneous modality profiles, whereas the FM embedding makes modality-specific distances more uniform.

Pareto coverage (\pcov) further reveals how much trade-off structure remains in the retrieved set. Even without the \emph{Pareto Gold} refinement reported later (\autoref{sec:res-isect}), the direct Pareto rerankers retain high coverage: Pareto (PCA) reaches 0.448 and Pareto (NN) reaches 0.440, above both scalar baselines (0.364--0.388) and the FM baseline (0.382). However, high \pcov\ is not limited to explicit Pareto selection: $k$NN [\indepemb] attains $\pcov = 0.417$, also exceeding the FM baseline. This suggests that \pcov\ is driven not only by Pareto selection, but also by whether the retrieval space preserves modality-specific geometry. When that geometry is preserved, non-dominated candidates remain visible even under standard nearest-neighbor retrieval. Scalar fusion and shared-space retrieval, in contrast, more often compress cross-modal trade-offs into a single ranking signal.

Together, the \cmb\ and \pcov\ results show that the main benefit of multi-objective retrieval is structural, not simply proximity-based. \paretofuse{} better preserves the non-dominated candidates that arise from uneven, modality-specific relevance. This makes the retrieved set less uniform, but more faithful to the cross-modal trade-offs that scientific retrieval must expose.

\subsection{Pareto Gold is compatible with the FM yet more selective}
\label{sec:res-isect}

\emph{Pareto Gold} substantially refines plain Pareto retrieval (\autoref{tab:main}): \rprox\ falls from 0.346 for Pareto (NN) to 0.249, while \pcov\ rises from 0.440 to 0.480, the highest value achieved. The $k$NN refinement step concentrates the Pareto front on candidates that remain close in the embedding space, improving average proximity while retaining non-dominated structure. The main trade-off appears in \cmb\ ($0.109 \rightarrow 0.044$), indicating that per-modality proximity becomes less uniform.

\begin{figure}[b]
\centering
\includegraphics[width=0.7\textwidth]{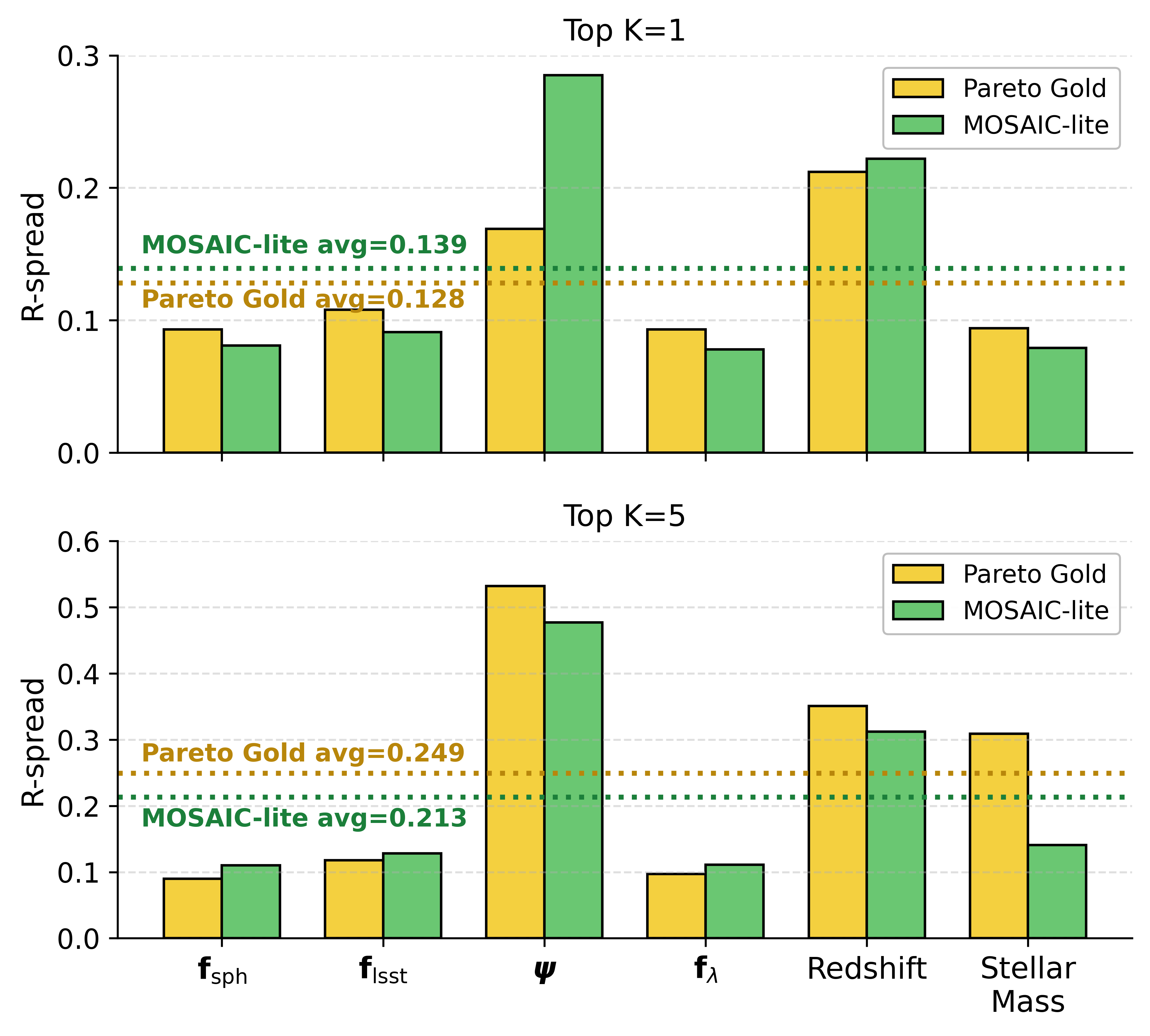}
\caption{Raw-data spread (\rspread) for \emph{Pareto Gold} and the FM baseline at $K{=}1$ and $K{=}5$. \rspread\ measures the average raw-space distance of the retrieved items to a given query; lower values indicate closer retrieved items in the original feature space. Dotted lines indicate per-method averages across the six modalities.}
\label{fig:raw-spread-k}
\end{figure}

We then examine whether this stricter modality-aware selection remains compatible with the similarity structure learned by the FM baseline. At matched retrieval depth $K$, \autoref{tab:gold-recall} shows that overlap is common but selective. Coverage rises from 0.556 at $K{=}1$ to 0.942 at $K{=}25$, so the two methods usually share at least one object as the retrieval set grows. However, \emph{Pareto Gold} recall falls from 0.556 to 0.110, meaning that most of its retrieved items are not recovered by the FM at larger depths. Thus, \emph{Pareto Gold} remains aligned with the broad FM neighborhood while prioritizing different candidates within it.

\begin{figure}[b]
\centering
\includegraphics[width=0.8\textwidth]{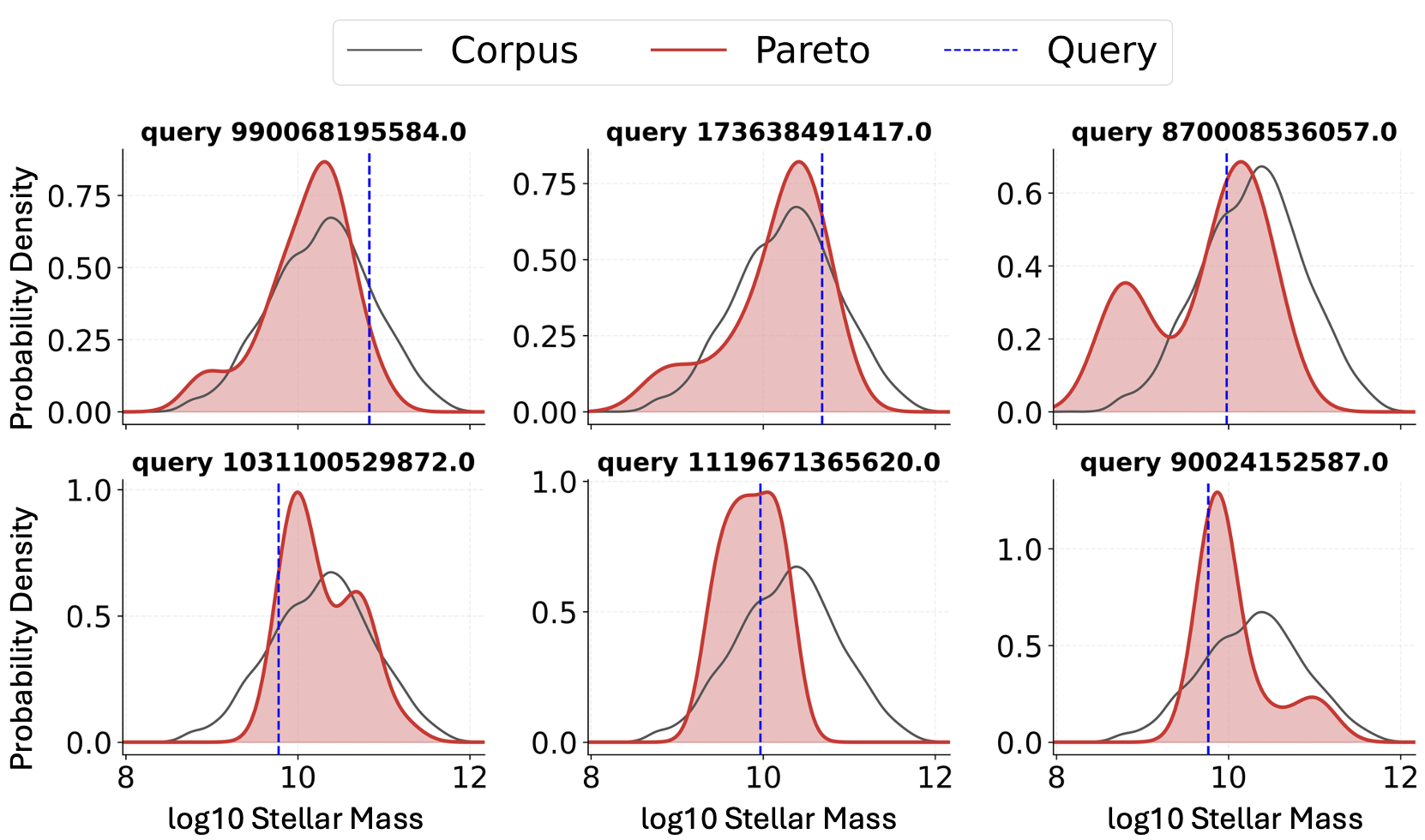}
\caption{Log stellar mass distributions for six example queries. Each panel shows the corpus density, the Pareto-front density, and the query value. The Pareto-front density is more concentrated near the query than the corpus background, indicating that non-dominated retrieval over observed modalities can align with a held-out scalar modality.}
\label{fig:pdf-stmass}
\end{figure}

\subsection{Pareto Gold improves physical fidelity}
\label{sec:res-spread}

Raw-data spread (\rspread) reveals a different ranking than embedding-space proximity. At $K{=}1$, \emph{Pareto Gold} achieves a lower average spread across the six modalities than the FM baseline (0.128 versus 0.139; \autoref{fig:raw-spread-k}), with the largest improvement on $\boldsymbol{\psi}$ (0.285 to 0.169, a 39\% relative reduction). This result is notable because the underlying Pareto front has roughly twice the embedding-space \rprox\ of the FM baseline (0.346 versus 0.153; \autoref{tab:main}). Thus, the method that is closest under the learned embedding metric is not necessarily the method that retrieves the closest neighbor in the original feature space. At $K{=}5$, the aggregate average spread slightly favors the FM baseline (0.213 versus 0.249), but \emph{Pareto Gold} remains closer on $\mathbf{f}_{\rm sph}$, $\mathbf{f}_{\rm lsst}$, and $\mathbf{f}_{\lambda}$, showing that its advantage persists for several modalities. These raw-space gains are noteworthy because Pareto retrieval uses only independently trained encoders, without cross-modal supervision or shared parameters.

The query-level examples provide a qualitative illustration of the same effect. In \autoref{fig:pdf-stmass}, the stellar-mass densities of Pareto-front items are more concentrated around the query value than the corpus background. These examples suggest that non-domination across observed modalities can serve as a useful proxy for similarity along physical attributes not directly used in retrieval.

\section{Discussion}
\label{sec:discussion}

We frame retrieval over heterogeneous modalities as a multi-objective optimization problem. Rather than projecting modality-specific embeddings into a single learned geometry, \paretofuse{} preserves each modality as a separate objective and uses Pareto dominance to retain candidates that remain competitive across all of them. The empirical question is whether this late fusion view can recover scientifically meaningful neighbors that are not surfaced by a unified FM embedding. Our results show that the FM baseline provides a strong global representation, whereas Pareto-based late fusion exposes trade-offs among the underlying physical measurements.

The main implication is that embedding-space proximity and raw-space fidelity are not equivalent. FM embeddings produce compact neighborhoods in its own learned space, as reflected by its lower \rprox. However, this intrinsic compactness does not always translate into proximity in the original data spaces. At the head of the ranking, \emph{Pareto Gold} retrieves neighbors with lower raw-space spread than the FM baseline on average. This indicates that a unified embedding can be highly effective as a semantic compression of the data while still obscuring modality-specific variation that remains scientifically meaningful.

This discrepancy reflects a fundamental difference in how the two retrieval strategies define similarity. A unified FM embedding induces one composite distance, so local neighborhoods are determined by the geometry learned across all modalities jointly. This is valuable when the goal is a reusable cross-modal representation or when downstream tasks benefit from smooth alignment across observed and unobserved modalities. Pareto-based retrieval imposes a different criterion: a candidate need not be nearest under a single aggregate distance, but must remain competitive across the preserved modality axes. This makes the ranking less compact in embedding space, but more explicit about which modalities support a neighbor relationship and where the trade-offs occur. In scientific settings, that distinction is important because similarity is often not a scalar property; two galaxies may be close in photometry, star-formation history, or emitted spectrum for different physical reasons.

The overlap analyses in \autoref{sec:res-isect} support this interpretation. Pareto-based retrieval and FM-based $k$NN do not select unrelated regions of the corpus; instead, they often identify a similar candidate neighborhood and then impose different local orderings within it. The shared candidates tend to be high-fidelity cases, while the method-specific candidates reveal where the two geometries diverge. This suggests a complementary division of labor: the FM embedding is well suited for broad candidate localization, whereas Pareto-based reranking can refine that neighborhood using modality-aware criteria. In this sense, the proposed method should not be understood as replacing FM retrieval, but as making explicit the modality-wise structure that a single learned distance may suppress.

Our results should not be read as a general critique of multimodal FMs. They instead identify a retrieval regime in which Pareto-based late fusion is most useful: modality-specific distance must remain inspectable and labeled data are too limited for task-specific reranking. In this setting, Pareto dominance provides a principled way to refine retrieval without reducing all modalities to a single shared representation. For discovery-oriented retrieval, where the meaning of scientific similarity depends on the measurements being compared, preserving this trade-off structure is a methodological advantage rather than a secondary diagnostic.

\section{Limitations}
\label{sec:limits}

This study has three main limitations. First, the evaluation considers a single FM baseline, MOSAIC-lite, and defines Pareto dominance using one-dimensional neural and PCA projections. The observed gap between embedding-space proximity and raw-space fidelity may therefore depend on the encoder, training objective, and projection quality. For PCA, the low explained variance ratio—the fraction of variance retained by the first principal component—observed for modalities such as $\boldsymbol{\psi}$ suggests that the projection may discard meaningful structure and weaken the dominance relation. Higher-dimensional or more expressive projections could preserve more information, but would increase the cost of Pareto reranking.

Second, dominance-based retrieval becomes more expensive as the corpus size and number of objectives grow. The four-modality setting studied here is tractable, but larger deployments may require approximate dominance pruning or other scalable retrieval mechanisms. Although our methods can be parallelized across corpus partitions and combined through a final ranking stage, we do not yet evaluate such an implementation with a larger number of modalities.

Third, our evaluation metrics partially reflect the retrieval objective in the dataset. Embedding-space metrics such as \rprox\ are useful for analyzing geometry, but they may not fully represent scientific relevance in the original measurements. This mismatch helps explain why our methods appear weaker under aggregated embedding-space scores while retrieving neighbors with stronger physical fidelity. Future work should develop evaluation criteria that better align embedding-space structure with raw-space agreement.

\section{Conclusion}
\label{sec:conclusion}

We introduced \paretofuse, a multi-objective late fusion framework for scientific modalities, and evaluated it on a galaxy retrieval task against an FM embedding baseline. Rather than collapsing all modalities into a single shared representation, the framework preserves modality-specific distance signals and combines them at retrieval time through Pareto dominance. The \emph{Pareto Gold} refinement further converts the resulting non-dominated set into a fixed-size ranked shortlist while retaining the multi-objective structure of the candidate pool.

Our results indicate that FM-based $k$NN retrieval and Pareto-based retrieval over independent embeddings capture different notions of similarity. The FM provides stronger proximity in embedding space but produces rankings that are more homogeneous across modalities. In contrast, \paretofuse{} retains modality-specific trade-offs and identifies neighbors that are closer to the query in the raw data space, particularly for informative modalities whose structure may be compressed in a shared embedding space. The two approaches therefore do not retrieve from entirely disjoint regions of the corpus; rather, they operate within similar neighborhoods while imposing different orderings on the candidates.

These findings suggest that late fusion of independent embeddings offers a practical complement to FM-based retrieval for scientific datasets with heterogeneous modality coverage. \paretofuse{} does not require retraining a unified model and provides a lightweight way to surface astrophysically coherent candidates whose relevance may be diluted in a single-objective ranking.

\bmhead{Acknowledgements}
This material is based upon work supported by the U.S. Department of Energy (DOE), Office of Science, Office of Advanced Scientific Computing Research, including the TIDES project and Argonne Leadership Computing Facility at Argonne National Laboratory, under Contract No. DE-AC02-06CH11357. 




\section*{Declarations}

The authors declare no conflict of interest.



\begin{appendices}
\section{MOSAIC and MOSAIC-lite foundation models for Cosmology}
\label{secA1}

MOSAIC~\cite{xia2025multi}, short for Multimodal Observation-Simulation Integration for Cosmology, is a multimodal foundation model (FM) designed to connect simulation-derived galaxy properties with synthetic observational quantities. It uses an encoder-only Transformer, inspired by BERT~\cite{devlin2019bert}, to operate on heterogeneous galaxy modalities, including scalar quantities such as redshift, halo mass, and stellar mass, as well as vector modalities such as $\mathbf{f}_{\rm lsst}$, $\mathbf{f}_{\rm sph}$, $\boldsymbol{\psi}$, and $\mathbf{f}_{\lambda}$. MOSAIC was trained on a subset of galaxies derived from the Last Journey simulation~\cite{heitmann2021last}, where galaxy properties and synthetic observables are assigned to dark-matter structures through galaxy--halo connection and stellar-population modeling.

\begin{figure}[!b]
    \centering
    \includegraphics[width=\linewidth]{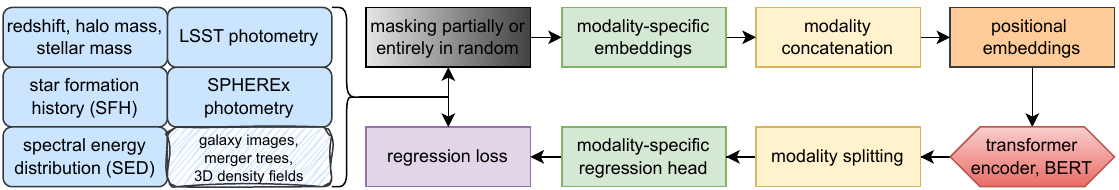}
    \caption{
    Schematic of the MOSAIC architecture. Each modality is masked and projected into modality-specific embeddings, concatenated into a shared token sequence, combined with positional embeddings, and processed by an encoder-only Transformer. The output sequence is split into modality-specific segments and decoded with dedicated regression heads. MOSAIC-lite follows the same shared-embedding principle and is used in this work as a lightweight multimodal FM baseline for retrieval.
    }
    \label{fig:mosaic_architecture}
\end{figure}

MOSAIC represents each modality as a sequence of continuous-valued tokens rather than using discrete tokenization. Scalar quantities are grouped into a ``scalar'' vector, while vector modalities naturally define token sequences over their sampled dimensions. As shown in \autoref{fig:mosaic_architecture}, each token is projected into a shared hidden dimension using a modality-specific linear embedding layer. The embedded modality sequences are concatenated along the sequence dimension, augmented with positional embeddings, and passed through a shared Transformer encoder. The output sequence is then split back into modality-specific segments and decoded using separate regression heads. This design allows the model to learn a shared latent representation while preserving modality-specific structure.

The model is trained using a masked-regression objective. During training, each modality may be partially or fully masked, simulating missing observations or unmeasured galaxy properties. The model is then asked to reconstruct both masked and unmasked values across modalities. This training strategy supports both in-modality reconstruction, such as recovering missing segments of $\mathbf{f}_{\lambda}$ or $\boldsymbol{\psi}$, and cross-modality prediction, such as predicting redshift, stellar mass, or spectra from photometric inputs. In the original MOSAIC study, combining $\mathbf{f}_{\rm lsst}$ and $\mathbf{f}_{\rm sph}$ improved redshift estimation relative to $\mathbf{f}_{\rm lsst}$ alone, while combining late-time $\boldsymbol{\psi}$ with $\mathbf{f}_{\rm lsst}$ improved stellar-mass inference relative to early-time $\boldsymbol{\psi}$ with $\mathbf{f}_{\rm lsst}$.

In this work, we use MOSAIC-lite as an FM baseline. Because the full MOSAIC model was trained on a substantially larger dataset, using it directly would conflate retrieval performance with differences in data scale and model capacity. We therefore train MOSAIC-lite on the same 20{,}000 galaxy samples as our evaluation, using the same multimodal inputs described above together with scalar modalities grouped into a scalar vector. MOSAIC-lite follows the same general principle as MOSAIC: it learns a shared embedding space through joint multimodal training. For the experiments in this paper, MOSAIC-lite uses an encoder-only Transformer with hidden size 384, 8 hidden layers, 12 attention heads, intermediate size 3072, and dropout probability 0.1 for both hidden states and attention weights. Retrieval with MOSAIC-lite is performed by nearest-neighbor search in its shared 384-dimensional embedding space.





\end{appendices}


\section*{Tool Use}

ChatGPT was used to improve the grammar and clarity of the manuscript. All suggested revisions were reviewed by the authors, who take full responsibility for the final content. The research ideas, methods, and results are the authors' own.

\bibliography{sn-bibliography}

\end{document}